\documentclass[lettersize,journal]{IEEEtran}
\usepackage{amsmath,amsfonts}
\usepackage{algorithmic}
\usepackage{algorithm}
\usepackage{array}
\usepackage[caption=false,font=normalsize,labelfont=sf,textfont=sf]{subfig}
\usepackage{textcomp}
\usepackage{stfloats}
\usepackage{xurl}
\usepackage{xcolor}
\usepackage{url}
\usepackage{verbatim}
\usepackage{graphicx}
\usepackage{cite}
\begin{document}

\title{GAN-Blot: A Controllable Structure–Style Synthesis Benchmark for Western Blot Forensics}

\author{Hao-Chiang Shao$^{\dagger}$,~\IEEEmembership{Senior Member,~IEEE}, 
Fong-Yi Lin, Te-An Chien~\IEEEmembership{Student Member,~IEEE}, TianYu Chen, Da-Jhong Chen~\IEEEmembership{Student Member,~IEEE}
\thanks{$^\dagger$H.-C. Shao is the corresponding author.}
\thanks{H.-C. Shao, F.-Y. Lin, D.-J. Chen, and T.-A. Chien 
are with the Institute of Data Science and Information Computing, National Chung Hsing University, Taiwan.}
\thanks{T. Chen is a Ph.D. student in the Department of Electrical Engineering, National Taiwan University of Science and Technology, Taiwan.}
}

\maketitle

\begin{abstract}
Western blot (WB) images are widely used as key evidence in biomedical research. Recent scientific misconduct cases reveal that WB imagery is increasingly fabricated, making WB forensics a major concern for research integrity. However, while the progress of forensic detection techniques often relies on advances in forgery-generation techniques, the development of WB forensic techniques has been hindered by the lack of standardized appearance attribute definitions, image datasets, and controllable generation frameworks for WB imagery. To address this limitation, we present a controllable WB image synthesis framework, named GAN-Blot, for generating realistic synthetic WB images. We introduce a formulation that decomposes a WB image into a structure component and a style-reference component, enabling independent control over local protein-band geometry and the global visual appearance of a synthetic WB image. GAN-Blot integrates a dual-path autoencoding design with several style-alignment loss terms to enable implicit control over structure–style synthesis without predefined semantic appearance attributes. We further contribute a synthetic WB dataset containing more than 46K images and propose four evaluation protocols for controllable WB synthesis. Extensive experiments show that GAN-Blot can generate WB images with high fidelity in both protein-band structure and visual style. Under blind inspection, the generated images can fool domain experts and are not reliably distinguished from authentic WB images by existing detectors and screening platforms. These results demonstrate their utility as challenging controlled cases for validating and developing WB forensic methods.

\end{abstract}

\begin{IEEEkeywords}
Scientific image forensics, Western blot imaging, Controllable image synthesis, AI-generated images
\end{IEEEkeywords}

\section{Introduction}
\label{sec01:intro}

\begin{figure}
    \centering
    \includegraphics[width=0.48\textwidth]{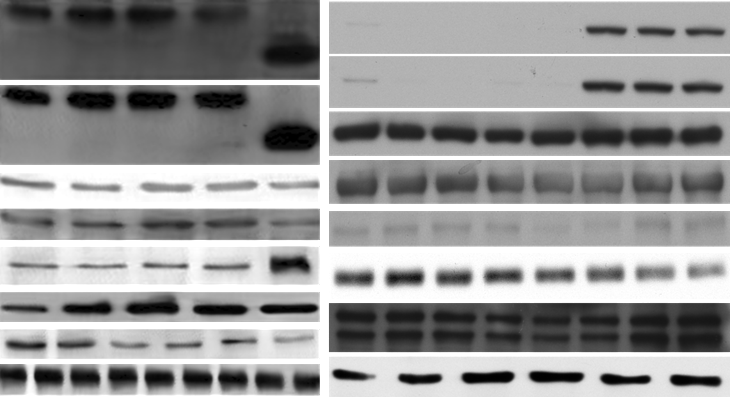}
    \caption{\textbf{Authentic or Synthetic?} Western blot (WB) images are digital recordings of molecular signal deposition rather than visual scenes. Consequently, faint bands, smooth backgrounds, and blurred band boundaries can arise in authentic WB imaging and should not be interpreted as evidence of image synthesis. Applying natural-image priors to WB images can lead to incorrect forensic judgments, motivating a generation-first benchmark for scientific-image forensics. Readers are invited to identify each of the 16 rows as authentic or synthetic. The answer is provided on the next page.
    }
    \label{fig:fig01_teaser}
\end{figure}

\IEEEPARstart{S}{cientific} image falsification has become one of the most prevalent forms of research misconduct in biomedical research, driving growing interest in scientific-image forensics. Recently, even major scientific publishers, such as Springer Nature~\cite{SpringerNature}, Science~\cite{Science}, and American Society for Microbiology~\cite{chaturvedi2025asm}, have incorporated image-screening platforms, e.g., \textbf{Proofig}~\cite{Proofig} and \textbf{ImageTwin}~\cite{Imagetwin}, into their editorial workflows for the pre-publication forensic screening of submitted figures, demonstrating that scientific integrity has become increasingly dependent on advances in image forensic techniques. However, most existing forensic techniques were originally developed for natural images, whose visual characteristics differ substantially from those of scientific images. This domain gap motivates the following forensic question illustrated in Fig.~\ref{fig:fig01_teaser}: \textit{Can state-of-the-art image forensic detectors, or even experienced forensic specialists, reliably distinguish synthesized Western blot images from authentic ones?} This paper addresses this question by developing \textbf{GAN-Blot}, a controllable WB image synthesis framework, together with a synthetic WB benchmark dataset to support the validation and development of WB forensic methods.

This question is directly relevant to Western blot (WB) images, which are used to quantify specific proteins and have become common targets of falsification. The WB workflow is labor-intensive and often takes several days to complete. Specifically, proteins are first separated by molecular weight through electrophoresis, then transferred onto a membrane, visualized as horizontal bands through antibody-based detection, and finally recorded by photography or scanning~\cite{mahmood2012western,laemmli1970cleavage,towbin1979electrophoretic}. Given their evidential role and the time required to produce WB results, WB images have become among the most frequently manipulated image modalities, as reflected in major research misconduct cases~\cite{MCCOOK2018Cancer,piller2022blots,kaiser2023should}. Recent advances in generative models and image-editing tools further introduce AI-related threats to academic integrity~\cite{davydiuk2025rising}. These issues have motivated research into WB image forensics.

However, although AI-generated image detectors have recently been developed for natural images or face images~\cite{ojha2023towards,koutlis2024leveraging,tan2024rethinking,yan2024effort,liu2024forgery}, detection techniques specifically designed for scientific-image falsification, especially WB-image falsification, remain limited. Existing WB forensic methods have primarily focused on conventional manipulations, including splicing~\cite{forensicdroplets,shao2018unveiling}, copy-move forgery, duplication, and image reuse~\cite{imachek,shao2024detecting,shao2025copy}. More importantly, advances in modern generative AI have introduced a different forensic scenario, in which an entire WB image can be fabricated without conducting experiments. 
As illustrated in Fig.~\ref{fig:fig01_teaser}, AI-generated trace detection is not necessarily equivalent to scientific-image falsification detection. For WB image forensics, the key forensic concern is whether an image contains patterns that cannot be produced by authentic WB experiments, rather than patterns that seem to be AI-generated artifacts but can arise in authentic WB experiments\footnote{\textbf{Answer to Fig.~\ref{fig:fig01_teaser}}: All images in the left column are synthesized by our proposed GAN-Blot, whereas all images in the right column are authentic laboratory Western blot images. Fig.~\ref{fig:fig01_teaser} illustrates how easily natural-image intuition can lead to incorrect judgments on scientific images.}. Although recent studies have begun investigating AI-generated WB content~\cite{mandelli2022forensic,manjunath2024localization,cardenuto2024explainable}, standardized generation benchmarks and systematic evaluation protocols are still lacking. These limitations make a standardized generation benchmark and a controllable WB image generator essential for advancing AI-related WB forensics, since new detectors cannot be systematically evaluated without them.

\begin{figure}[!t]
    \centering
    \includegraphics[width=0.48\textwidth]{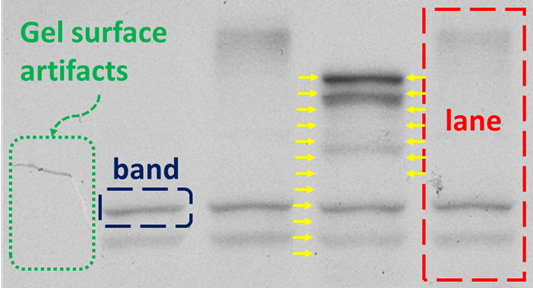}
    \caption{
    \textbf{Domain-specific visual attributes of a Western blot (WB) image.} Key WB attributes include i) the numbers of lanes, bands, and rows; ii) foreground band morphology; iii) lane-trace contour and intensity; and iv) background characteristics such as intensity, noise patterns, and gel surface artifacts/textures. A \textbf{lane} denotes a vertical electrophoretic channel where a sample is separated by molecular weight. A \textbf{band} is a horizontal intensity region within a lane representing proteins of similar molecular weight. The \textbf{trace}, indicated by yellow arrows, refers to the faint vertical boundary or intensity profile outlining a lane. The lack of standardized definitions for these attributes makes controllable WB synthesis difficult and motivates our structure--style formulation. 
    }
    \label{fig01:illustrate}
\end{figure}

A reliable forgery generator and a standardized evaluation benchmark can accelerate progress in forensic detection techniques~\cite{verdoliva2020media,tolosana2020deepfakes}. Representative examples include FaceForensics++~\cite{Rossler2019FaceForensics++}, Celeb-DF~\cite{Li2020CelebDF}, and DFDC~\cite{Dolhansky2020DFDC}, all of which provide realistic forged samples for systematic detector evaluation. Following this generation-first paradigm, AI-related WB forensics likewise requires a controllable WB synthesis benchmark for forensic validation. However, developing such a benchmark is challenging because WB images lack standardized semantic descriptions and appearance annotations. As shown in Fig.~\ref{fig01:illustrate}, a WB image is characterized by bands, lane traces, background intensity, and gel/membrane artifacts, but these domain-specific factors do not correspond to the semantic attributes commonly used in natural-image generation. Consequently, recent text-guided or instruction-guided generation models~\cite{zhang2023adding,brooks2023instructpix2pix} cannot easily provide precise control over WB foreground structures and background appearance while maintaining consistency with WB experimental characteristics. This limitation motivates our structure--style formulation for WB synthesis. 

More specifically, we formulate WB synthesis as a dual-component conditional generation problem, in which protein-band \textit{structure} and gel/membrane \textit{style} are specified by separate inputs and controlled independently. Based on this formulation, we propose GAN-Blot, a controllable WB image generator for synthesizing realistic WB images for scientific-image forensic research. GAN-Blot generates WB images with controllable band structures, lane patterns, intensity distributions, and gel/membrane styles while preserving plausible visual characteristics of WB experiments. We further construct a synthetic WB dataset and evaluation protocols to assess the generated images in terms of visual realism, controllability, expert assessment, and forensic utility. 

Our contributions are thus threefold. \\
\noindent $\bullet$\textbf{Structure--style formulation for controllable WB synthesis.} We formulate controllable WB synthesis as a structure--style conditional generation problem, in which protein-band geometry is specified by a band-contour mask and gel/membrane appearance is specified by a style-reference image. This formulation enables independent control over local band structures and global visual appearance without relying on predefined semantic appearance attributes. \\
\noindent $\bullet$\textbf{Dual-path conditional autoencoding for combinatorial structure--style learning.}
We design a dual-path conditional autoencoding training procedure for GAN-Blot. Instead of learning only from fixed structure--style pairs, GAN-Blot learns from randomly recombined conditions, where $N$ mask patterns and $M$ style patterns can form up to $N \times M$ structure--style compositions. The self-supervision path further anchors these recombined conditions to realistic WB appearance by reconstructing real WB images from their extracted masks and style codes. This design enables GAN-Blot to learn complex structure--style synthesis from limited WB images without predefined appearance labels or textual descriptions.
\noindent $\bullet$\textbf{Synthetic WB benchmark and forensic evaluation.}
We construct a synthetic WB benchmark dataset generated by GAN-Blot and establish four evaluation protocols covering controllability, target-conditioned resynthesis, expert-based visual assessment, and forensic evaluation under existing detectors. Our experiments show that GAN-Blot generates WB images with high fidelity in both protein-band structure and visual style. The generated images can fool domain experts under blind inspection and are not reliably distinguished from authentic laboratory WB images by existing detectors and publisher-used screening platforms, demonstrating their utility as challenging controlled cases for validating and developing WB forensic methods. 

%
%
\section{Related Work}
\label{sec02:review}

Recent AI-generated image detectors have been developed mainly for natural images or face images~\cite{ojha2023towards,koutlis2024leveraging,tan2024rethinking,yan2024effort,liu2024forgery,zhu2023gendet}. In parallel, existing WB forensic methods have primarily addressed conventional scientific-image falsification, including splicing, copy-move, duplication, and image reuse~\cite{forensicdroplets,shao2018unveiling,imachek,shao2024detecting,shao2025copy}. Recent studies have further begun investigating AI-generated WB images~\cite{mandelli2022forensic,manjunath2024localization,cardenuto2024explainable}. These works provide important forensic baselines and evaluation targets. In contrast, this paper focuses on controllable WB image synthesis for constructing synthetic WB images that can be used to validate WB forensic methods.

Although recent advances in image generation have made controllable synthesis possible, their direct application to WB image synthesis remains challenging due to the absence of standardized appearance attribute definitions and structured annotations. Recent text- or instruction-guided diffusion models achieve controllability through semantic prompts or spatial conditioning signals~\cite{zhang2023adding,brooks2023instructpix2pix,li2023gligen}. These methods rely on textual descriptions or grounded annotations to regulate content and appearance. In WB imagery, however, structural components such as lanes and bands do not correspond to standardized semantic entities, limiting the effectiveness of prompt-based or spatial conditioning for precise structural control. Similarly, attribute-supervised controllable synthesis methods often require predefined labels, such as identity, pose, or age attributes~\cite{shao2021dotfan,huang2022age,zhang2020face,wang2020cross,huang2023towards}. Since WB imagery lacks such standardized appearance annotations, these strategies are difficult to apply directly.

Existing WB synthesis attempts remain limited. Current approaches primarily rely on direct image-to-image (I2I) translation~\cite{mandelli2022forensic}, without explicit mechanisms to regulate WB-specific structures such as lane configurations and band geometry. Also, no clear task formulation or benchmark dataset exists for evaluating controllable WB synthesis or its forensic utility. Related controllability designs in other domains, such as DreamPose~\cite{karras2023dreampose}, Animate Anyone~\cite{hu2024animate}, and Men \textit{et al.}~\cite{men2022unpaired}, guide synthesis using external conditioning or reconstruction constraints to preserve structural content while transferring appearance. These designs suggest that controllable synthesis can be achieved by separating \textit{structure} and \textit{style} without predefined attribute taxonomies.

Motivated by these works, we formulate WB synthesis as a structure–style conditional generation problem. Compared with direct I2I translation~\cite{mandelli2022forensic}, our formulation enables stronger structural controllability and provides a foundation for constructing standardized WB synthesis benchmarks.

\section{Method}
\label{sec03:method}
\begin{figure*}[!t]
    \centering
    \includegraphics[width=0.88\textwidth]{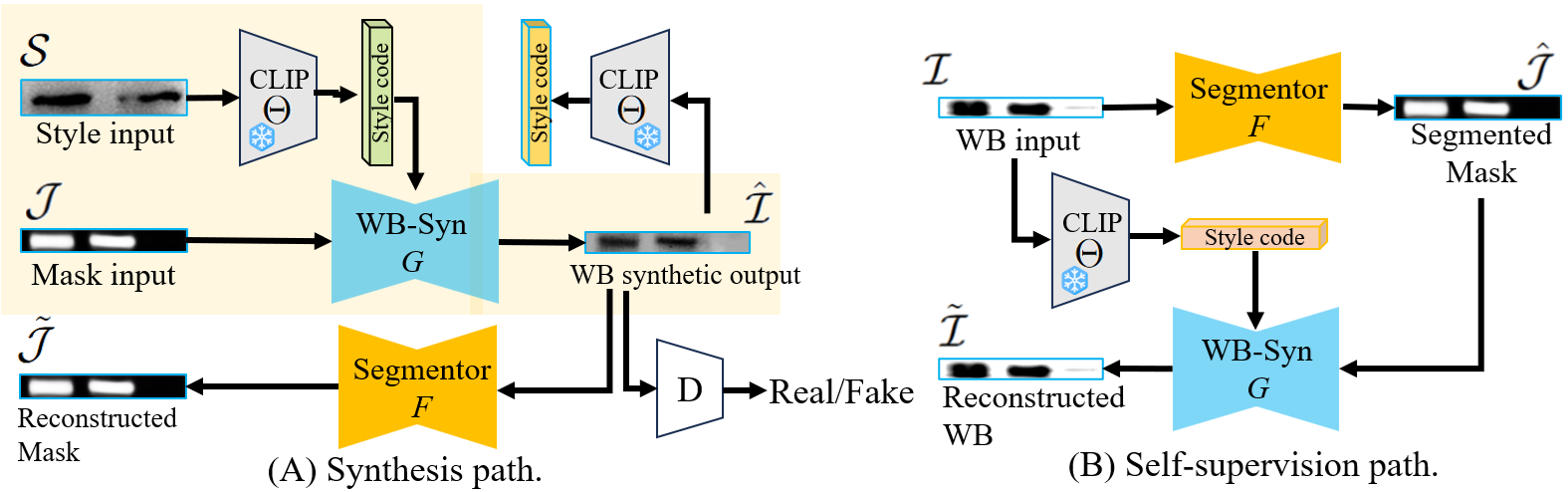}
    \caption{\textbf{Training framework of GAN-Blot.} The model is trained with triplet training samples, each consisting of a randomly sampled style reference $\mathcal{S}$, a WB image $\mathcal{I}$, and its band-contour mask $\mathcal{J}$ in a dual-path manner. While the synthesis path learns structure–style synthesis, the self-supervision path enforces accurate reconstruction of $\mathcal{I}$ and consistency of its corresponding band mask, facilitating stable learning. 
    }
    \label{fig:framework}
\end{figure*}

\subsection{Overview}
\label{subsec:301}
GAN-Blot formulates WB synthesis as a dual-component conditional generation problem. Given a band-contour mask $\mathcal{J}$ representing protein-band structure and a style-reference image $\mathcal{S}$ providing gel/membrane appearance, the goal is to synthesize a realistic WB image $\hat{\mathcal{I}}$ that preserves the protein-band geometry specified by $\mathcal{J}$ while adopting the appearance characteristics of $\mathcal{S}$. In this formulation, the structure component describes local band locations, shapes, and lane layouts, whereas the style component captures global WB appearance, including background intensity, noise patterns, gel/membrane textures, and lane-trace clarity. 
This structure--style formulation differs from existing I2I strategies that translate a binary protein-band mask directly into a WB image~\cite{mandelli2022forensic,manjunath2024localization}. Such binary-to-grayscale mapping is inherently ill-posed because a single band mask can correspond to many plausible gel/membrane appearances. Since appearance is not explicitly conditioned, these methods learn appearance variation implicitly, often resulting in limited appearance diversity and weak background control. By introducing an independent style-reference input, GAN-Blot provides explicit appearance guidance during synthesis, thereby improving appearance controllability and diversity without requiring predefined appearance labels or textual descriptions.

As illustrated in Fig.~\ref{fig:framework}, GAN-Blot consists of a WB-Synthesizer $G$, a protein-band segmentor $F$, a pretrained CLIP visual encoder $\Theta$ used as the style encoder, and a discriminator $D$. The synthesizer $G$ takes the band-contour mask $\mathcal{J}$ and the style code $\mathbf{f}_\mathcal{S}=\Theta(\mathcal{S})$ as inputs to generate the synthetic WB image $\hat{\mathcal{I}}=G(\mathcal{J},\mathbf{f}_\mathcal{S})$. During training, GAN-Blot uses triplet samples $(\mathcal{I},\mathcal{J},\mathcal{S})$, where $\mathcal{I}$ is a real WB image, $\mathcal{J}$ is its band-contour mask, and $\mathcal{S}$ is a randomly sampled style-reference image. This random pairing prevents the model from learning only fixed mask--image pairs and instead exposes it to recombined structure--style conditions, where $N$ mask patterns and $M$ style patterns can form up to $N \times M$ structure--style compositions.

To learn from these recombined conditions, GAN-Blot adopts a dual-path conditional autoencoding scheme in which $G$ and $F$ are jointly optimized. The synthesis path, shown in Fig.~\ref{fig:framework}(A), learns conditional WB generation by enforcing structural consistency with the input mask and style alignment with the reference image. The self-supervision path, shown in Fig.~\ref{fig:framework}(B), reconstructs real WB inputs from their extracted protein-band masks and style codes, anchoring the learned structure--style conditions to photorealistic WB appearance. Formally, the dual-path training encourages the following two consistency relations:
\begin{equation}
F(G(\mathcal{J}, \mathbf{f}_\mathcal{S})) \simeq \mathcal{J}
\quad \mbox{and} \quad
G(F(\mathcal{I}), \mathbf{f}_{\mathcal{I}}) \simeq \mathcal{I},
\end{equation}
where $\mathbf{f}_\mathcal{S}=\Theta(\mathcal{S})$ and $\mathbf{f}_\mathcal{I}=\Theta(\mathcal{I})$ denote the style codes extracted from the style-reference image $\mathcal{S}$ and the real WB image $\mathcal{I}$, respectively.

The first relation encourages the synthesized image to preserve the input protein-band mask, while the second relation encourages the generator to reconstruct a real WB image from its extracted protein-band mask and style code. Under this training scheme, GAN-Blot learns to use the mask $\mathcal{J}$ and the style-reference $\mathcal{S}$ as separate conditions for WB synthesis. During inference, only the synthesizer $G(\cdot)$ is used to generate $\hat{\mathcal{I}}=G(\mathcal{J},\Theta(\mathcal{S}))$, as highlighted in the light yellow region in Fig.~\ref{fig:framework}. The following subsections describe the network components and loss functions in detail.

Finally, note that we adopt StyleGAN2 rather than a diffusion-based model as the backbone of our synthesizer because our StyleGAN2-based backbone provides a direct mechanism for injecting reference-style information into the generation process and thus facilitates structure--style recombination. In contrast, training diffusion-based models usually requires more domain-specific data to learn a stable target distribution~\cite{zhang2025training}, making convergence and controllability more challenging given limited WB training samples. Their iterative sampling process also increases the cost of generating large numbers of structure--style combinations~\cite{ma2025efficient}.

\subsection{Subnetworks}

\textbf{GAN-Blot} consists of three primary subnetworks---namely, a WB-Synthesizer $G$, a protein-band segmentor $F$, and a style encoder $\Theta$---along with a discriminator $D$ for adversarial training. The architecture of each component is described as follows.

\begin{figure*}[!t]
    \centering
    \includegraphics[width=0.85\textwidth]{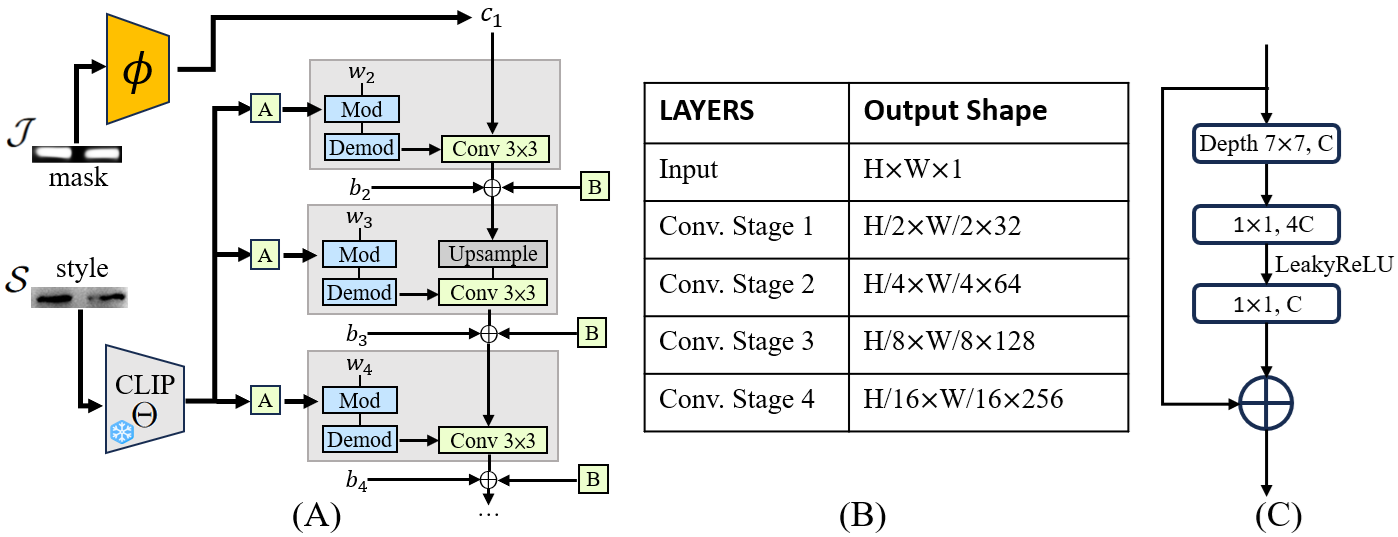}
    \caption{\textbf{Architecture of the WB-Synthesizer in GAN-Blot.} (A) The WB-Synthesizer extends StyleGAN2 by integrating a band-contour encoder $\phi$ and a style encoder $\Theta$. (B) Architecture of $\phi$. (C) Modified ConvNeXt block within $\phi$.}
    \label{fig:WBsynArch}
\end{figure*}

\noindent $\bullet$\textbf{WB-Synthesizer.} The WB-Synthesizer is built upon the StyleGAN2 architecture~\cite{karras2020analyzing}, with two key modifications, as illustrated in Fig.~\ref{fig:WBsynArch}(A). First, the fixed learnable constant feature tensor $c_1$ in the original StyleGAN2 is replaced by a feature tensor extracted from the protein-band mask $\mathcal{J}$ via a learnable encoder $\phi$, \textit{i.e.}, $c_1 = \phi(\mathcal{J})$. Second, the latent style input is replaced with the style feature extracted from the style-reference image $\mathcal{S}$ by the style encoder $\Theta$. These modifications enable the synthesizer to generate photorealistic WB images conditioned jointly on the protein-band mask and the style-reference image, while keeping the synthesized image at the same resolution as the input mask. 

\noindent $\bullet$\textbf{Synthesizer Encoder and Discriminator.} The encoder $\phi$ of the synthesizer is a ConvNeXt-based~\cite{liu2022convnet} feature extractor, as shown in Fig.~\ref{fig:WBsynArch}(B). It consists of four convolutional stages, each structured as \textit{Conv}$1{\times}1$$\rightarrow$\textit{ConvNeXt Block}$\rightarrow$\textit{ConvNeXt Block}$\rightarrow$\textit{Downsampling by 2}. 
In our implementation of the \textit{ConvNeXt Block}, layer normalization is removed, and LeakyReLU is employed as the activation function, as illustrated in Fig.~\ref{fig:WBsynArch}(C). For adversarial training, a PatchGAN discriminator~\cite{zhu2017unpaired} is employed to promote local realism in the synthesized WB images.

\noindent $\bullet$\textbf{Segmentor.} As illustrated in Fig.~\ref{fig:framework}, the purpose of the protein-band segmentor $F$ is not merely to perform band segmentation, but to ensure that $\hat{\mathcal{J}} = F(\mathcal{I})$ serves as a 2D structural code for autoencoding the real WB image $\mathcal{I}$. To prevent direct leakage of contour information through skip connections, $F$ adopts an autoencoder architecture \emph{without} horizontal skip connections, unlike common U-Net-based designs. This constraint forces protein-band shape information to be encoded through bottleneck features rather than transferred directly. 

\noindent $\bullet$\textbf{Style Encoder.} The style encoder $\Theta$ is a pretrained CLIP image encoder. Following Xiao~et~al.~\cite{xiao2021early}, convolutional backbones are more effective than transformer backbones for capturing structural features; hence, we adopt the ResNet-50 variant of CLIP instead of the ViT version. Although CLIP is trained on natural images rather than WB imagery, its image embedding provides a compact representation that can be used as a style code. In GAN-Blot, this code serves as a reference-guided appearance descriptor, enabling style alignment without predefined WB appearance labels or textual descriptions. 

Together, these subnetworks enable mask-guided structure control and reference-guided appearance control, which are further enforced by the loss functions described below.

\subsection{Loss Functions}

\noindent $\bullet$\textbf{Losses from Synthesis Path.} 
The synthesis path is responsible for learning structure-style synthesis. For this path, we define four loss terms, including (i) the band-shape loss $\mathcal{L}^G_\mathrm{band}$ between the input mask $\mathcal{J}$ and the reconstructed mask $\tilde{\mathcal{J}}$, (ii) the histogram loss $\mathcal{L}_\mathrm{hist}$ between the synthesized output $\hat{\mathcal{I}}$ and the style input $\mathcal{S}$, (iii) the style-code loss $\mathcal{L}_\mathrm{code}$ to enforce similarity in style embeddings, and (iv) the style loss $\mathcal{L}_\mathrm{style}$ to preserve texture-level consistency. These four losses are described below. 

To preserve the protein-band structure throughout the synthesis process, we enforce the consistency between the estimated protein-band mask of the synthesized output, \textit{i.e.}, $\tilde{\mathcal{J}}=F(\hat{\mathcal{I}})$, and the input mask $\mathcal{J}$. The band-shape loss of $G$ is defined as: 
\begin{equation}
    \mathcal{L}^G_\mathrm{band} = \ell_\mathrm{BCE}( \mathcal{J}, \tilde{\mathcal{J}}) \mbox{,} \label{eq:bce1}
\end{equation}
where $\ell_\mathrm{BCE}$ denotes the binary cross-entropy loss.  

Next, to align the synthesized image $\hat{\mathcal{I}}$ with the style reference $\mathcal{S}$, we introduce three complementary style-related loss terms.    
The first is the \textbf{histogram loss}, which aligns the global brightness and contrast between $\mathcal{S}$ and $\hat{\mathcal{I}}$. Since the Earth Mover’s Distance between two histograms is equivalent to the $L_1$ distance between their cumulative distribution functions (CDFs)~\cite{werman1985distance,avi2023differentiable}, we define:    
\begin{eqnarray}    
\mathcal{L}_\mathrm{hist} &=& \| \mathcal{H}\big(\mathcal{S}\odot (\mathbf{1}-M_\mathcal{S}) \big) - \mathcal{H}\big({\hat{\mathcal{I}}}\odot (\mathbf{1}-M_\mathcal{I})\big) \|_1 \nonumber \\
&& 
+\| \mathcal{H}\big(\mathcal{S}\odot M_\mathcal{S}\big) - \mathcal{H}\big({\hat{\mathcal{I}}}\odot M_\mathcal{I}\big) \|_1 
\mbox{.}
\label{eq:histogramloss}
\end{eqnarray}
Here, $M_\mathcal{S}$ denotes the protein-band mask of the style reference  $\mathcal{S}$, $\odot$ indicates element-wise multiplication, $M_\mathcal{I}$ is set to the input mask  $\mathcal{J}$, and $\mathcal{H}(\cdot)$ denotes the histogram CDF. This loss consists of two terms: one measuring the background-histogram discrepancy and the other measuring the discrepancy within the foreground protein-band region.

\noindent The second is the \textbf{style-code loss}, which enforces feature-level similarity between the CLIP-encoded representations of $\mathcal{S}$ and $\hat{\mathcal{I}}$:
\begin{equation}
    \mathcal{L}_\mathrm{code} = 1 - \cos \big( \Theta(\mathcal{S}), \Theta(\hat{\mathcal{I}}) \big) \mbox{.}
    \label{eq:cos}
\end{equation}
The third is the \textbf{style-consistency loss}, which enforces visual texture consistency between $\mathcal{S}$ and $\hat{\mathcal{I}}$. Since texture style is mainly captured by high-level feature activations~\cite{gatys2016image}, we compute this term using the perceptual loss $\ell_\mathrm{perc}$~\cite{johnson2016perceptual} derived from the fourth convolutional layer:
\begin{equation}
    \mathcal{L}_\mathrm{style} = \ell_\mathrm{perc}^{j=4}(\mathcal{S}\odot (\mathbf{1}-M_\mathcal{S}), \hat{\mathcal{I}}\odot (\mathbf{1}-M_\mathcal{I})) \mbox{.}
\end{equation}
Here, $\mathcal{L}_\mathrm{style}$ is computed only on the background region, and average pooling is applied to align the spatial dimensions of the tensors.

\noindent $\bullet$\textbf{Losses from Self-supervision Path.}  The self-supervision path ensures accurate reconstruction of each WB input and structural consistency of its corresponding segmented protein-band mask. It contributes two loss terms, including the band-shape loss $\mathcal{L}^F_\mathrm{band}$ between the input mask $\mathcal{J}$ and the segmented mask $\hat{\mathcal{J}}$, as well as the reconstruction loss $\mathcal{L}_\mathrm{rec}$ between the WB input $\mathcal{I}$ and its reconstruction $\tilde{\mathcal{I}}$.

These two losses are defined as: 
\begin{eqnarray}
    \mathcal{L}^F_\mathrm{band} &=& \ell_\mathrm{BCE}( \mathcal{J}, \hat{\mathcal{J}}) \mbox{, and} \label{eq:bce2}
    \\
    \mathcal{L}_\mathrm{rec} &=& \| \mathcal{I} - \tilde{\mathcal{I}} \|_1 \mbox{.} \label{eq:recloss}
\end{eqnarray} 
With the aid of the reconstruction loss $\mathcal{L}_\mathrm{rec}$, the segmented band mask $\hat{\mathcal{J}}$ is encouraged to serve as a 2D structural code for autoencoding the input WB image, thereby supporting mask-guided control of protein-band structures.

\noindent $\bullet$\textbf{GAN Loss.} 
\indent Because the PatchGAN discriminator violates the input independence assumption of WGAN-GP~\cite{gulrajani2017improved} due to overlapping patches and therefore causes instability during training~\cite{shao2023retina}, we adopt the relativistic pairing GAN (RpGAN)~\cite{huang2024gan} loss instead:
\begin{equation}
    \mathcal{L}_\mathrm{GAN} = \mathbb{E} \big\{ f\big( \mathbf{D}(\mathbf{G}(\mathcal{J}, \mathbf{f}_\mathcal{S}))  - \mathbf{D}(\mathcal{I}) \big)\big\} \mbox{,}
\end{equation}
where $f(t) = -\log(1+e^{-t})$ follows the RpGAN formulation, and $\mathbb{E}\{\cdot\}$ denotes expectation.

\subsection{Total Loss and Model Optimization}

The total loss functions for the WB synthesizer $G$, discriminator $D$, and segmentor $F$ are defined as
\begin{eqnarray}
    \mathcal{L}^G_\mathrm{total} 
    &=& \mathcal{L}_\mathrm{GAN} + 10\mathcal{L}_\mathrm{rec} + 10\mathcal{L}^G_\mathrm{band} + 10\mathcal{L}_\mathrm{hist} \nonumber\\
    && + 0.1\mathcal{L}_\mathrm{code} + 0.1\mathcal{L}_\mathrm{style} \mbox{,} \nonumber \\
    \mathcal{L}^D_\mathrm{total} 
    &=& \mathcal{L}_\mathrm{GAN} + \mathcal{R}_1 + \mathcal{R}_2 \mbox{,} \nonumber \\
    \mathcal{L}^F_\mathrm{total} 
    &=& \mathcal{L}^F_\mathrm{band} + 10\mathcal{L}_\mathrm{rec} + 10\mathcal{L}^G_\mathrm{band} \mbox{.}
\end{eqnarray}
Here, $\mathcal{R}_1$ and $\mathcal{R}_2$ denote the regularization terms introduced in~\cite{huang2024gan} to improve convergence stability. $\mathcal{R}_1$ penalizes the gradient norm of $D$ on real data, and $\mathcal{R}_2$ penalizes the gradient norm of $D$ on generated data. In each training iteration, $D$ is first updated, followed by the joint update of $G$ and $F$. Additionally, because the loss terms were integrated into the model progressively (\textit{i.e.}, one at a time during development), the weighting coefficients for $\mathcal{L}_\mathrm{rec}$, $\mathcal{L}^G_\mathrm{band}$ and $\mathcal{L}^F_\mathrm{band}$, $\mathcal{L}_\mathrm{hist}$, $\mathcal{L}_\mathrm{code}$, and $\mathcal{L}_\mathrm{style}$ were empirically tuned based on the visual quality of the synthesized WB images obtained during training. Each newly introduced loss term was tuned while keeping the previously selected coefficients fixed, allowing its coefficient to be adjusted independently. 

Finally, each training batch contains one  sample, including a mask image $\mathcal{J}$, its paired WB image $\mathcal{I}$, and a randomly selected WB image serving as the style reference $\mathcal{S}$. The model was trained for \textbf{approximately 1000K iterations}. 
%
%
\section{Experiment Results}
\label{sec04:exp}

We evaluate GAN-Blot from four aspects, including controllability, target-conditioned resynthesis, forensic evaluation, and expert assessment. For model development, we collect 4939 non-AI-generated Western blot (WB) images with band-contour masks for training and testing GAN-Blot. These source images are used only for model development and will not be publicly released. Instead, we construct \textbf{WesternBlot46K} as a forensic research dataset consisting of (i) 45,861 fully synthetic WB images generated by GAN-Blot and (ii) an authentic laboratory subset that contains 151 cropped authentic laboratory WB images and 195 full resolution authentic WB photos provided by domain collaborators in 2025. \textbf{WesternBlot46k} will be released for academic research.

We compare GAN-Blot with representative GAN-based models, text- or instruction-guided generative models, including ControlNet~\cite{zhang2023adding} and InstructPix2Pix~\cite{brooks2023instructpix2pix}, and image-to-image (I2I) translation models under the same mask-conditioning setting. For forensic evaluation, we test GAN-Blot outputs using publicly released AI-generated image detectors, WB-specific synthetic-image detectors, and image-screening platforms used in publisher editorial workflows, including Proofig~\cite{Proofig} and ImageTwin~\cite{Imagetwin}. Domain-expert assessment is further conducted to evaluate whether the synthesized WB images are visually plausible under human inspection. An ablation study analyzes the contribution of each loss component.

\subsection{Dataset Preparation}

For model development, we collect 4939 
non-AI-generated Western blot (WB) images from 84 
retracted or publicly questioned papers. Among them, 4000 
images are used for training---thus forming up to 4K$\times$4K=16M 
random combinations---and 939 
images are used for testing. In addition, we include 151 authentic cropped WB images provided by a molecular biology laboratory, forming a testing set of 939+151= 1090 
images. The publication-cropped images are collected from electronic publications reported on Retraction Watch or discussed on PubPeer. To avoid redistributing copyrighted source figures, these publication-cropped images are used only for model development and are not publicly released. For transparency and reproducibility, we will provide a source-paper list containing paper titles, publication venues, years, and public case information, together with our cropping and preprocessing protocol in our project page.

The collected WB images span diverse biomedical domains (e.g., cancer biology, molecular biology, immunology, \textit{etc.}), multiple journal tiers (e.g. \textit{PNAS}, \textit{JBC}, \textit{Scientific Reports}, \textit{etc.}), and more than two decades from 2001 to 2019. This diversity introduces substantial variability in experimental protocols, imaging devices, acquisition conditions, and visual characteristics, reducing potential domain- or device-specific bias. During preprocessing, non-image information such as panel labels, captions, surrounding text, and laboratory-identifying annotations is removed. Each WB image $\mathcal{I}$ is associated with a protein-band mask $\mathcal{J}$ annotation, obtained via intensity-based thresholding followed by manual inspection. Masks with evident contour errors are excluded from the training set, whereas masks with minor isolated noise or slightly irregular boundaries are retained as noisy but acceptable annotations. This decision reflects the ambiguous boundaries of weak protein bands in real WB images and avoids retaining only high-contrast, overly clean band contours. 

Using the trained GAN-Blot model, we generate a fully synthetic subset consisting of 45,861 WB images. The released research dataset will contain the GAN-Blot-generated images and the authentic laboratory subset, while the publication-cropped training images will not be redistributed. The generated samples and authentic laboratory subset will be released for academic, non-commercial research with registered access. Existing public WB synthesis datasets~\cite{mandelli2022forensic,manjunath2024localization} contain only small re-synthesized image patches and are therefore less suitable for controllable full-image WB synthesis.

\subsection{Experimental Protocols and Compared Methods}
%

We evaluate GAN-Blot using four protocols that examine structural controllability, appearance controllability, target-specific reconstruction fidelity, and forensic evaluation.
The first two protocols evaluate whether GAN-Blot can separately control WB structure and appearance. The third protocol evaluates whether the proposed training design can reconstruct a target WB image under designated structural and appearance conditions. The fourth protocol evaluates whether existing forensic detectors and image-screening platforms can reliably identify realistic AI-generated WB images.

\noindent \textbf{P1. Structural controllability.}
This protocol evaluates whether the synthesized WB image preserves the protein-band layout specified by the input band-contour mask $\mathcal{J}$. Given the same style-reference image $\mathcal{S}$, we synthesize WB images from different band-contour masks and examine whether the resulting band locations, shapes, and lane layouts follow the corresponding masks.

\noindent \textbf{P2. Appearance controllability.}
This protocol evaluates whether the synthesized WB image follows the appearance characteristics of the style-reference image $\mathcal{S}$. Given the same band-contour mask $\mathcal{J}$, we synthesize WB images using different style-reference images and examine whether background intensity, contrast, lane-trace visibility, noise patterns, and gel/membrane textures vary according to the reference input.

\noindent \textbf{P3. Target-specific reconstruction fidelity.}
This protocol evaluates whether a model can reconstruct a target WB image when its band-contour mask and appearance reference are provided. For each testing image $\mathcal{I}$, we use its band-contour mask as the structural input and the image itself as the style reference. The reconstructed image is evaluated using appearance-based metrics, including MSE, PSNR, SSIM, and LPIPS, together with band-structure consistency measured by IoU and style similarity measured by perceptual loss.

\noindent \textbf{P4. Forensic evaluation under existing detectors.}
This protocol evaluates the forensic utility of a WB synthesizer by testing whether existing forensic tools can reliably distinguish synthesized WB images from non-synthetic WB images. We consider three groups of detectors and screening tools. The first group includes publicly released AI-generated image detectors developed mainly for natural or face images~\cite{ojha2023towards,koutlis2024leveraging,tan2024rethinking,yan2024effort,liu2024forgery,zhu2023gendet}. The second group includes WB-specific synthetic-image detection methods~\cite{manjunath2024localization,cardenuto2024explainable}. The third group includes image-screening platforms used in publisher editorial workflows, including Proofig~\cite{Proofig} and ImageTwin~\cite{Imagetwin}. In this protocol, images generated by the evaluated WB synthesizer are treated as synthetic samples, whereas authentic laboratory WB images and non-AI-generated publication-cropped WB images are treated as non-synthetic samples. The results are used to evaluate whether the generated WB images can be reliably distinguished from non-synthetic WB images by existing forensic methods. 

\begin{figure*}[!t]
    \centering    \includegraphics[width=0.95\textwidth]{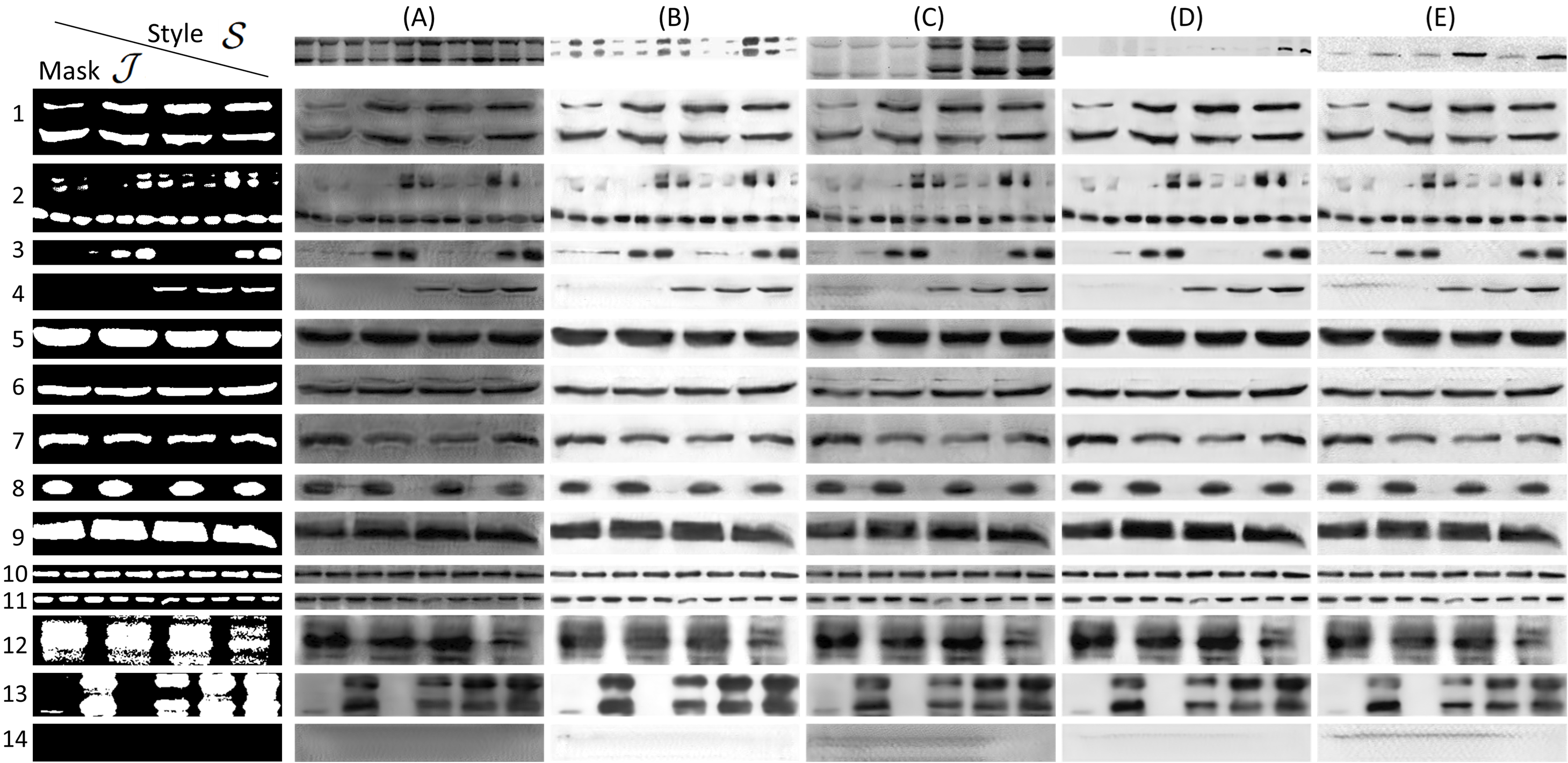}
    \vspace{-0.3cm}
    \caption{\textbf{Examples of WB images generated by GAN-Blot.} The leftmost column shows the input protein-band mask  $\mathcal{J}$, and the top row shows the input style reference $\mathcal{S}$. 
    GAN-Blot is capable of generating WB images with independently controllable band structures and appearance styles.} 
    \label{fig:showall}
\end{figure*}
\begin{figure*}[!t]
    \centering
    \includegraphics[width=0.95\textwidth]{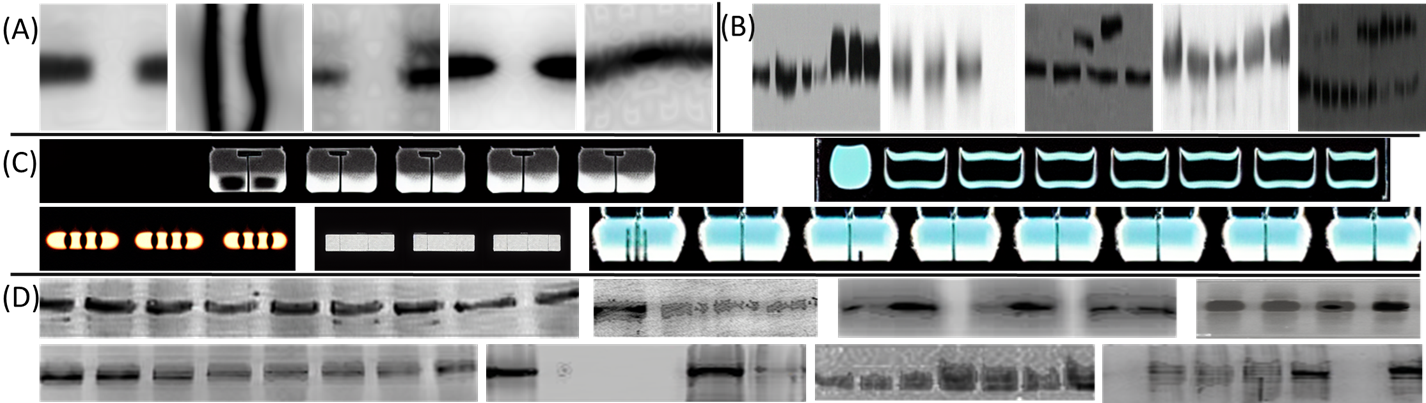} \vspace{-0.3cm} 
    \caption{\textbf{WB images synthesized by (A) StyleGAN3~\cite{karras2021alias}, (B) ADM~\cite{dhariwal2021diffusion}, (C) InstructPix2Pix~\cite{brooks2023instructpix2pix}, and (D) ControlNet~\cite{zhang2023adding}.}  Results from StyleGAN3 and ADM show that random noise inputs cannot achieve structure or style control. 
    In contrast, results from InstructPix2Pix and ControlNet suggest that text- or instruction-based methods cannot reliably synthesize WB images, because WB appearance lacks well-defined semantic attribute descriptions and these approaches rely primarily on textual instructions without explicit style-reference inputs for controlling WB appearance.
    } 
    \label{fig:gan_and_diffusion}
\end{figure*}

For controllable synthesis evaluation, we compare GAN-Blot with representative GAN-based, diffusion-based, and image-to-image translation baselines. These include StyleGAN3~\cite{karras2021alias}, ADM~\cite{dhariwal2021diffusion}, ControlNet~\cite{zhang2023adding}, InstructPix2Pix~\cite{brooks2023instructpix2pix}, and CycleGAN~\cite{zhu2017unpaired}. For target-specific reconstruction in \textbf{P3}, we further implement two conditioned StyleGAN2-based baselines, namely, StyleGAN2-J and StyleGAN2-JS. StyleGAN2-J uses the synthesizer encoder $\phi$ and is conditioned only on $\mathcal{J}$, while StyleGAN2-JS additionally uses the style encoder $\Theta$ and is conditioned on both $\mathcal{J}$ and $\mathcal{S}$. These two baselines share the same generator backbone as GAN-Blot and are used to separate the effect of architectural conditioning from the proposed dual-path training and loss design.

In addition to these four protocols, we conduct an expert-based visual assessment to examine whether synthesized WB images are visually plausible under human inspection.

\subsection{Controllability of WB Synthesis}

We first evaluate whether GAN-Blot can separately control protein-band structures and gel/membrane appearance characteristics, \textit{i.e.}, \textbf{P1} and \textbf{P2}. As shown in Fig.~\ref{fig:showall}, GAN-Blot synthesizes WB images using different combinations of band-contour masks and style-reference images. When the same style-reference image is paired with different masks, the generated WB images preserve the band locations, shapes, lane layouts, and empty-lane regions specified by the corresponding masks. This result indicates that the band-contour mask $\mathcal{J}$ provides effective structural control over the synthesized WB image. In particular, when a mask contains no protein-band regions (\textit{i.e.}, the mask shown in the 14th row of Fig.~\ref{fig:showall}), GAN-Blot does not hallucinate unintended bands, suggesting that the generated foreground structure is primarily guided by the input mask. 

Moreover, when the same band-contour mask is paired with different style-reference images, GAN-Blot preserves the band structure while adapting the overall appearance to the reference image. The generated images reflect changes in background intensity, contrast, lane-trace visibility, noise patterns, and gel/membrane textures according to the style reference. These results demonstrate that the style-reference image $\mathcal{S}$ provides effective appearance control without changing the specified protein-band structure. Taken together, the results in Fig.~\ref{fig:showall} show that GAN-Blot supports mask-guided structure control and reference-guided appearance synthesis through separate inputs.

\begin{table*}[t]
    \centering
    \small
    \caption{\textbf{Performance comparison and ablation study via WB re-synthesis.} (A) Performance comparison among network architectures and (B) Loss-term ablation. All values are computed on images whose intensities are normalized to the [0,1] range } \vspace{-0.2cm}
    \begin{tabular}{|l|c|c|c|c||c||c|}
    \hline
       \multicolumn{7}{|c|}{(A) Module Ablation: Performance Comparison among Different Model Architectures}\\
        \hline
        Model  & MSE$\downarrow$ & PSNR$\uparrow$ & SSIM$\uparrow$ & LPIPS$\downarrow$ & IoU ($\tilde{\mathcal{J}}$)$\uparrow$ & Perc. Loss$\downarrow$\\
       \hline
       GAN-Blot & .0047$\pm$.0039 & 24.57$\pm$3.55 
       & .8322$\pm$.0969 & .1131$\pm$.0814 & .9978$\pm$.0188& .1101$\pm$.0972\\
       StyleGAN2-JS & .0250$\pm$.0230 & 17.17$\pm$2.97 & .6894$\pm$.1042 & .2835$\pm$.1108 & .8694$\pm$.1720 & .0661$\pm$.1537 \\
       StyleGAN2-J & .0518$\pm$.0419 & 14.11$\pm$3.40 & .6036$\pm$.1693 & .4518$\pm$.1588 & .4456$\pm$.1476& .3562$\pm$.4278\\
       CycleGAN & .1214$\pm$.0807 & 9.91$\pm$2.53 & .2792$\pm$.1814 & .5837$\pm$.1434 & .0130$\pm$.0198 & .3117$\pm$.3117\\
       \hline
    \hline
    \multicolumn{7}{|c|}{(B) Loss-term Ablation: Impact of Each Loss}\\
    \hline
       Loss  & MSE$\downarrow$ & PSNR$\uparrow$ & SSIM$\uparrow$ & LPIPS$\downarrow$ & IoU ($\tilde{\mathcal{J}}$)$\uparrow$ & Perc. Loss$\downarrow$\\
       \hline
       Full Model & .0047$\pm$.0039 & 24.57$\pm$3.55 & .8322$\pm$.0969 &  .1131$\pm$.0814 & .9978$\pm$.0188 & .1101$\pm$.0972 \\
       w/o $\mathcal{L}_\mathrm{hist}$  & .0238$\pm$.0223 & 17.73$\pm$3.67 & .7633$\pm$.1121 & .1669$\pm$.0932 & .9568$\pm$.0564 & .0766$\pm$.0606\\
       
       w/o $\mathcal{L}_\mathrm{style}$  & .0047$\pm$.0038 & 24.51$\pm$3.48 & .8314$\pm$.0974 & .1126$\pm$.0842 & .9711$\pm$.0236 & .1202$\pm$.1048\\       
       
       
       w/o $\mathcal{L}_\mathrm{code}$  & .0048$\pm$.0040 & 24.45$\pm$3.55 & .8274$\pm$.0975 & .1172$\pm$.0833 & .9704$\pm$.0238 & .1074$\pm$.0938\\
       
       w/o $\mathcal{L}^G_\mathrm{band}$  & .0057$\pm$.0045 & 23.76$\pm$3.60 & .8106$\pm$.1039 & .1227$\pm$.0828 & .9000$\pm$.0744 & .1111$\pm$.0967\\


       w/o
       $\mathcal{L}_\mathrm{rec}$  & .0053$\pm$.0045 & 24.06$\pm$3.52 & .8140$\pm$.1014 & .1211$\pm$.0866 & .9640$\pm$.0275 & .1346$\pm$.1239\\
       
       w/o 
       $\mathcal{L}^F_\mathrm{band}$  & .0050$\pm$.0039 & 24.20$\pm$3.38 & .8209$\pm$.0992 & .1203$\pm$.0862 & .9450$\pm$.0538 & .1135$\pm$.0966\\

       \hline
    \end{tabular}
    \label{tab:ablationstudy}
\end{table*}

We further compare GAN-Blot with representative generative baselines, as shown in Fig.~\ref{fig:gan_and_diffusion}. Noise-driven GAN and diffusion models can synthesize grayscale images with WB-like local patterns, but they lack explicit structural conditioning and therefore cannot guarantee correspondence between the synthesized bands and a user-specified mask. Text-guided or instruction-guided models, such as ControlNet~\cite{zhang2023adding} and InstructPix2Pix~\cite{brooks2023instructpix2pix}, provide a more flexible conditioning interface, but their outputs remain unstable for WB synthesis. This is because WB appearance characteristics, such as lane traces, background intensity, membrane texture, and band morphology, do not have standardized semantic descriptions comparable to those used in natural-image generation. As a result, these models often fail to simultaneously preserve the specified band structure and produce realistic WB background appearance.

These comparisons suggest that controllable WB synthesis cannot be reliably achieved by directly applying general-purpose generative models. GAN-Blot addresses this limitation by using a band mask $\mathcal{J}$ for structural guidance and a WB style-reference image $\mathcal{S}$ for appearance guidance, enabling controllable synthesis under WB-specific visual constraints.

\subsection{Target-Conditioned Resynthesis and Loss Ablation}
We next evaluate target-conditioned resynthesis under the \textbf{P3} protocol. Here, ``resynthesis'' refers to regenerating a WB image from its extracted protein-band mask and its own appearance reference. 
Given a testing WB image $\mathcal{I}$, we use its protein-band mask $\mathcal{J}$ as the structural input and the image itself as the style reference. The objective is to synthesize a WB image that matches the target image in protein-band structure and overall appearance characteristics. We compare GAN-Blot with CycleGAN and two conditioned StyleGAN2 baselines, namely StyleGAN2-J and StyleGAN2-JS. StyleGAN2-J is conditioned only on the protein-band mask, whereas StyleGAN2-JS is conditioned on both the protein-band mask and the style-reference image. Since StyleGAN2-JS shares the same conditioning inputs and generator backbone as GAN-Blot, this comparison helps isolate the effect of the proposed dual-path training and loss design.

Table~\ref{tab:ablationstudy}(A) reports the resynthesis fidelity. GAN-Blot achieves the best MSE, PSNR, SSIM, and LPIPS scores among the compared methods, indicating that the re-synthesized images are closer to the target WB images in both pixel-level fidelity and perceptual similarity. GAN-Blot also obtains the highest IoU between the input protein-band mask and the mask extracted from the re-synthesized image, showing that the specified protein-band structures are accurately preserved. In contrast, CycleGAN fails to maintain the target band layout because it does not explicitly use the protein-band mask as a structural condition. StyleGAN2-J improves structural control but lacks reference-guided appearance conditioning. StyleGAN2-JS further incorporates the style-reference image, but its resynthesis fidelity remains substantially lower than that of GAN-Blot. These results indicate that architectural conditioning alone is insufficient for target-conditioned WB resynthesis, and that the proposed training design supports the joint preservation of WB structure and appearance. 

\begin{figure*}[t]
    \centering
    \includegraphics[width=0.95\textwidth]{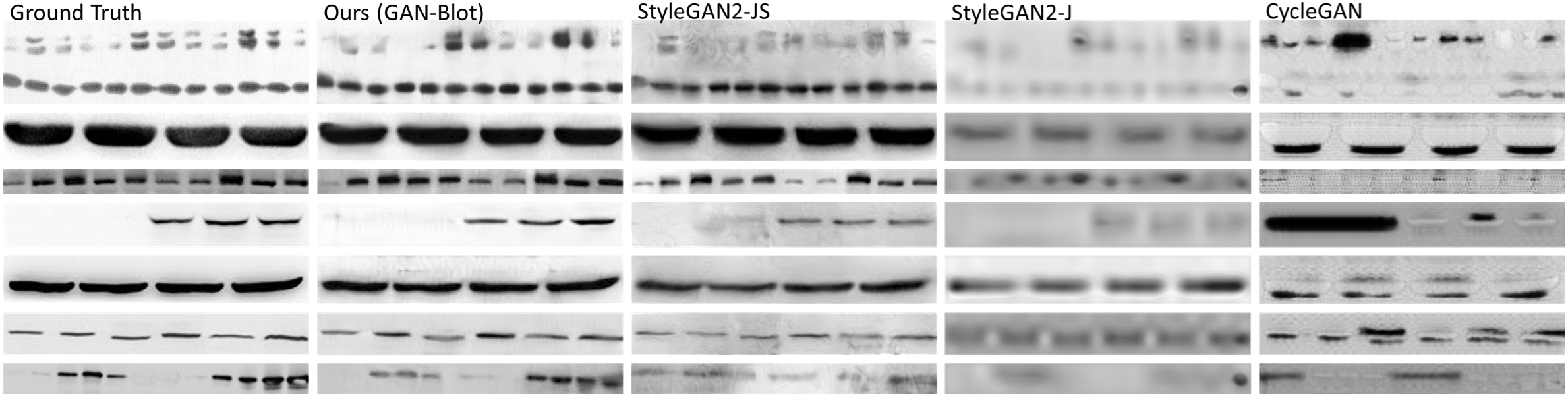} \vspace{-0.3cm}
    \caption{\textbf{Visualization of Protocol-P3 evaluation.} This figure presents the reconstruction results corresponding to Table~\ref{tab:ablationstudy}(A). Since StyleGAN2-JS and GAN-Blot’s WB-Synthesizer are architecturally identical, these examples clearly illustrate that the dual-path training scheme  and loss formulation adopted in GAN-Blot play a crucial role in WB image synthesis with controllable band contours and appearance style, enabling faithful synthesis of the desired target WB image.} \vspace{-0.1cm}
    \label{fig:show_other_model}
\end{figure*}

Fig.~\ref{fig:show_other_model} further supports this observation. GAN-Blot re-synthesizes the target WB images with faithful band locations, band shapes, lane patterns, background intensity, and gel/membrane texture. By contrast, the baseline methods either distort the band structures, produce inconsistent lane patterns, or fail to match the target background appearance. These examples are consistent with Table~\ref{tab:ablationstudy}(A) and show that GAN-Blot can use the protein-band mask and style-reference image as separate but complementary conditions for WB image synthesis.

\begin{table*}[!t]
    \centering
    \caption{\textbf{Expert visual evaluation of synthesized WB images.} Fooling Rate is computed under blind inspection. Structure Plausibility (Struct. Plaus.), Style Plausibility (Style Plaus.), and Realism Rate measure structural validity, stylistic validity, and overall realism, respectively. Parenthesized values indicate fooling rates from image-processing specialists without WB-domain expertise.} 
    \small
    \begin{tabular}{|c|c||c|c|c|c|}
    \hline
      Method   & \#Img. &  Fooling Rate & Struc. Plaus. & Style Plaus. & Realism Rate\\
      \hline
      GAN-Blot & 75 & 81.3\% (88.2\%) & 76.0\% & 80.0\% & 56.0\%\\
      StyleGAN2-JS & 50 & 74.0\% (57.0\%)&  61.3\% & 56.6\% & 38.7\%\\
      CycleGAN~\cite{zhu2017unpaired} & 50 & 10.0\% (22.0\%) & 7.3\% & 6.0\% & 0\%\\

      ControlNet~\cite{zhang2023adding} & 52 & 9.6\% (47.5\%) & 5.8\% & 7.7\% & 3.8\%\\
      InstructPix2Pix~\cite{brooks2023instructpix2pix} & 50 & 0 (0) & 0 & 0 & 0 \\
    \hline
    \end{tabular}
    \label{tab:expertexamination}
\end{table*}

We further conduct a loss ablation study to examine the contribution of each training objective. As shown in Table~\ref{tab:ablationstudy}(B), removing the histogram loss $\mathcal{L}_\mathrm{hist}$ causes a clear degradation in resynthesis fidelity, indicating that matching the intensity distribution is important for preserving global WB appearance, including background brightness and contrast. Removing the band-shape losses $\mathcal{L}^G_\mathrm{band}$ or $\mathcal{L}^F_\mathrm{band}$ reduces the structural consistency between the input mask and the generated image, confirming their importance for mask-guided protein-band preservation. Removing the style-code loss $\mathcal{L}_\mathrm{code}$ or the style-consistency loss $\mathcal{L}_\mathrm{style}$ weakens reference-guided appearance control, showing that both the CLIP-based style-code constraint and the background texture constraint contribute to appearance transfer from the reference image. Finally, removing the reconstruction loss $\mathcal{L}_\mathrm{rec}$ also degrades resynthesis fidelity, demonstrating that the self-supervision path helps the generator synthesize realistic WB images under self-derived structural and appearance conditions. Overall, the ablation results show that the proposed loss components jointly support structure preservation, appearance alignment, and target-conditioned WB resynthesis.

\subsection{Expert-based Visual Assessment}
We further conduct an expert-based visual assessment to examine whether the synthesized WB images are plausible under domain-expert inspection. The assessment involves three experts, including two researchers in molecular biology and one computer-vision researcher who has served on investigation committees for suspected scientific-image falsification. Each expert is asked to inspect WB images synthesized by different methods without knowing the generation method. We report the fooling rate, defined as the percentage of synthesized images that are not identified as synthetic by the experts.

In addition to the fooling rate, we further ask the experts to assess whether each synthesized image contains unrealistic protein-band structures or unrealistic gel/membrane appearance. Based on these annotations, we compute the structure plausibility, style plausibility, and realism rate. Structure plausibility measures the percentage of images without unrealistic protein-band layouts or band shapes. Style plausibility measures the percentage of images without unrealistic background appearance, lane traces, or gel/membrane textures. Realism rate measures the percentage of images judged as plausible in both structure and appearance.

Table~\ref{tab:expertexamination} reports the expert assessment results. GAN-Blot achieves the highest fooling rate, structure plausibility, style plausibility, and realism rate among the compared methods. In particular, the higher structure plausibility indicates that GAN-Blot better preserves WB-like protein-band layouts, while the higher style plausibility shows that the generated background appearance is more consistent with real WB images. By contrast, CycleGAN, ControlNet, and InstructPix2Pix frequently produce distorted band layouts, unrealistic background textures, or unstable lane patterns. StyleGAN2-JS obtains better results than these baselines, but still falls behind GAN-Blot, showing that the proposed dual-path training and loss design are important for synthesizing WB images with plausible structure and appearance.

We also include an additional assessment by 9 image-processing engineers without WB-domain expertise, with the averaged fooling rates reported in parentheses in Table~\ref{tab:expertexamination}. Their judgments differ from those of the domain experts. For StyleGAN2-JS, the fooling rate from image-processing engineers is lower than that from the domain experts, suggesting that its GAN-related artifacts are more recognizable to participants with computer-vision or image-processing experience. By contrast, CycleGAN and ControlNet receive much higher fooling rates from these engineers than from the domain experts, suggesting that participants without WB-domain knowledge may overlook unrealistic band structures, lane patterns, or gel/membrane appearance. These results show that synthesized WB images should be assessed from both biological and image-forensics perspectives. 

\begin{table*}[!t]
    \caption{Performance of existing AI-generated image detectors on GAN-Blot synthetic WB images and non-AI-generated WB images.}
    \centering
    \setlength{\tabcolsep}{4pt}
    \begin{tabular}{|c||c|c|c|c|c|c||c|c|c|c|c|c|c|c|c|}
    \hline
     & \multicolumn{6}{c||}{(A) GAN-Blot-Synthesized} & \multicolumn{9}{c|}{(B) Non-AI-generated} \\ \cline{2-16}

    & \multicolumn{3}{c|}{(A-1) Manually inspected} & \multicolumn{3}{c||}{(A-2) Randomly selected} & \multicolumn{3}{c|}{(B-1) Lab-Cropped} & \multicolumn{3}{c|}{(B-2) Lab-full} & \multicolumn{3}{c|}{(B-3) publication-cropped}\\ 

    Method & \multicolumn{3}{c|}{n=360} & \multicolumn{3}{c||}{n=1640} & \multicolumn{3}{c|}{n=151} & \multicolumn{3}{c|}{n=195} & \multicolumn{3}{c|}{n=1654}\\
    \cline{2-16}

    & TP & FN & TPR(\%) & TP & FN & TPR(\%) & TN & FP & FPR(\%) & TN & FP & FPR(\%) & TN & FP & FPR(\%) \\ \hline
    Effort~\cite{yan2024effort}  & 68 & 292 & 18.89    & 296 & 1344 & 18.05 & 99 & 52 & 34.44  & 133 & 62 & 31.79 & 1182 & 472 & 28.54 \\ \hline
       RINE~\cite{koutlis2024leveraging}  & 41 & 319 & 11.39 &  505 & 1135 & 30.79 &  151 & 0 & 0 & 175 & 20 & 10.26 & 1477 & 177 & 10.70  \\ \hline
       UnivFD~\cite{ojha2023towards}  & 3 & 357 & 0.83 & 415  & 1225 &  25.30 & 141  & 10 &  6.62 & 140 & 55 & 28.21 & 1289 & 365 & 22.07 \\ \hline
       FatFormer~\cite{liu2024forgery}  & 0 & 360 & 0    & 16 & 1624 &  0.98 & 151 & 0 &  0 & 191 & 4 & 2.05 & 1654 & 0 & 0 \\ \hline
       NPR~\cite{tan2024rethinking}  & 0 & 360   &  0   & 1 & 1639 & 0.06  & 151 & 0 &  0  & 195 & 0 & 0 & 1654 & 0 & 0 \\ \hline
       LNCLIP-DF~\cite{yermakov2026deepfake}  & 0 & 360  &  0   & 5 & 1635 &  0.30 & 151 & 0 &  0 & 195 & 0 & 0 & 1653 & 1 & 0.06 \\ \hline
    \end{tabular}
    \label{tab:ai_detect_5sets}
\end{table*}
\begin{figure*}[t]
    \centering
    \includegraphics[width=0.98\textwidth]{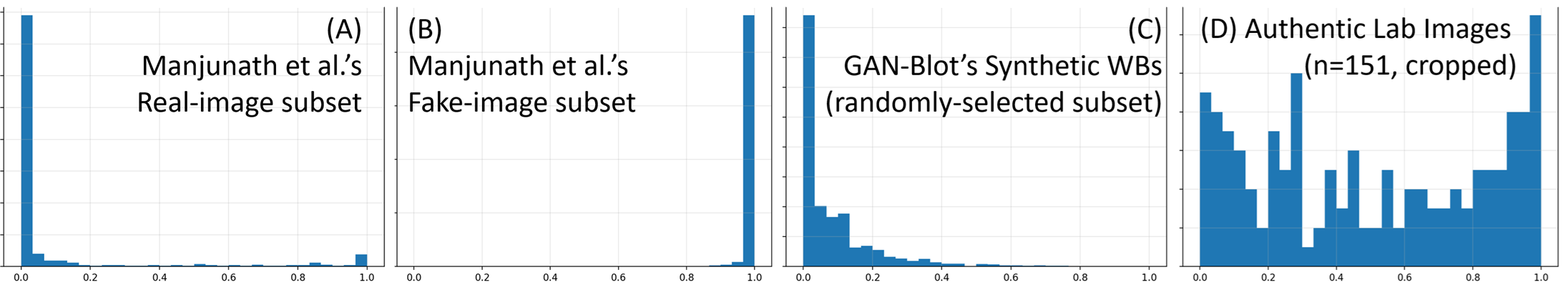}
    \vspace{-0.3cm}
    \caption{\textbf{Response distributions of Manjunath et al.’s method~\cite{manjunath2024localization}.} Histograms show the final softmax score distributions for \textbf{(A)} Manjunath et al.’s official real-image subset, \textbf{(B)} Manjunath et al.’s official fake-image subset, \textbf{(C)} the randomly selected GAN-Blot subset, and \textbf{(D)} the cropped authentic laboratory WB subset. The detector separates the official real and fake subsets, but does not reliably respond to WB images synthesized by GAN-Blot and lacks a stable response to authentic laboratory images. 
    }
    \label{fig08:4pics}
\end{figure*}

\subsection{Forensic Evaluation under Existing Detectors}

Here, we perform forensic evaluation under existing detectors, corresponding to the \textbf{P4} protocol. We first test publicly released AI-generated image detectors, including Effort~\cite{yan2024effort}, RINE~\cite{koutlis2024leveraging}, UnivFD~\cite{ojha2023towards}, FatFormer~\cite{liu2024forgery}, NPR~\cite{tan2024rethinking}, and LNCLIP-DF~\cite{yermakov2026deepfake}, which were developed mainly for natural or face images. To examine both synthetic-image detection and false-alarm behavior in the WB domain, we evaluate these detectors on five subsets, as summarized in Table~\ref{tab:ai_detect_5sets}. The two GAN-Blot image subsets include \textbf{(A-1)} a subset of GAN-Blot synthetic WB images, all of which were manually inspected for visual plausibility, and \textbf{(A-2)} a subset consisting of randomly selected GAN-Blot synthetic WB images. The three non-AI-generated WB subsets include \textbf{(B-1)} cropped authentic laboratory WB images, \textbf{(B-2)} full authentic laboratory WB photos without cropping, and \textbf{(B-3)} publication-cropped WB images sampled from non-AI-generated publication figures. This design allows us to evaluate whether existing detectors can identify GAN-Blot synthetic images while avoiding false alarms on laboratory and publication-style WB images.

\begin{figure*}[t]
    \centering
    \includegraphics[width=0.98\textwidth]{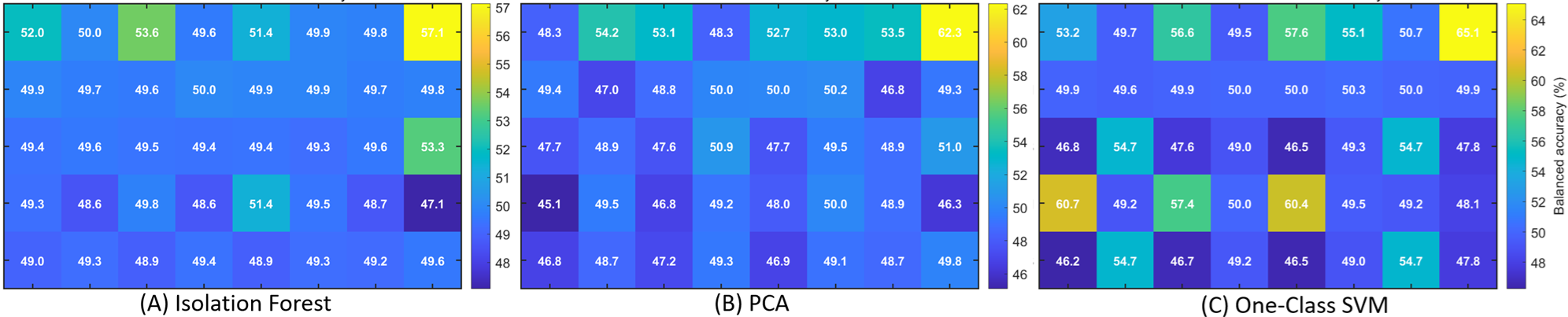}
    \vspace{-0.3cm}
    \caption{\textbf{Balanced accuracy of Cardenuto et al.’s method~\cite{cardenuto2024explainable} under official configurations.} Heat maps show the balanced accuracy (\%) of Cardenuto et al.’s 3$\times$5$\times$8=120 official configurations on GAN-Blot images, using \textbf{(A)} Isolation Forest, \textbf{(B)} PCA, and \textbf{(C)} One-Class SVM. Most configurations remain close to 50\%, indicating near-chance detection performance.
    }
    \label{fig09:3heats}
\end{figure*}

As shown in Table~\ref{tab:ai_detect_5sets}, existing AI-generated image detectors exhibit weak detection performance and unstable false-alarm behavior in the WB-image domain. On subset \textbf{(A-1)}, which consists of GAN-Blot synthetic WB images manually inspected for visual plausibility, all detectors achieve TPRs below 20\%. On the randomly selected subset \textbf{(A-2)}, most detectors still miss a large portion of the synthetic images, and only RINE reaches a TPR of about 30\%. Moreover, Effort and UnivFD produce non-negligible false positives on the non-AI-generated WB subsets in Table~\ref{tab:ai_detect_5sets}(B). By contrast, FatFormer, NPR, and LNCLIP-DF show nearly zero TPRs and FPRs, indicating that their low FPR values mainly come from a strong tendency to classify WB images as non-synthetic. Among the examined detectors, only RINE shows limited sensitivity to GAN-Blot synthetic WB images, but its detection capability remains insufficiently reliable. These results indicate that common AI-generated image detectors, although effective for natural-image or face-image forensics, do not adequately cover WB-image forensic scenarios in scientific-image integrity. Taken together, by providing a systematic framework for synthesizing realistic WB images and demonstrating that existing AI-generated image detectors are largely ineffective on GAN-Blot images, GAN-Blot offers controlled synthetic cases for advancing WB-specific synthesis and detection techniques.

After evaluating general AI-generated image detectors, we next test two WB-specific synthetic-image detectors proposed in recent studies. For Manjunath et al.’s method~\cite{manjunath2024localization}, we train it based on its official implementation and released dataset and test it in turn on (i) its official testing set, (ii) our randomly selected GAN-Blot subset, and (iii) our authentic laboratory image subset. Fig.~\ref{fig08:4pics} shows the histograms of the binary classification score output by the final softmax layer. On the official testing set, the trained model separates the real and fake subsets as reported by Manjunath et al.~\cite{manjunath2024localization}. However, when applied to GAN-Blot images, the distribution of the classification scores becomes close to that of the real-image subset, implying that the detector does not reliably respond to WB images synthesized by GAN-Blot. Fig.~\ref{fig08:4pics}(D) further shows a broad, nearly flat score distribution on authentic laboratory WB images, indicating that Manjunath et al.’s method also lacks a stable response to authentic laboratory images. 
As for Cardenuto et al.’s method~\cite{cardenuto2024explainable}, it aims to identify artifact-based features via combinations of 5 feature types, 8 residual filters, and 3 classifiers for AI-generated WB images. We run its official implementation without any modifications to derive the results. As shown in Fig.~\ref{fig09:3heats}, most combinations obtain low balanced accuracy (BACC) on GAN-Blot images. Among the 3$\times$5$\times$8=120 evaluated configurations, the median BACC is 49.51\%, only nine configurations exceed 55\%, and only four configurations exceed 60\%. These results indicate that Cardenuto et al.’s method mostly yields near-chance detection performance on GAN-Blot images under its official configurations.

We also conduct screening using the paid analysis services of publisher-used image-screening platforms, including ImageTwin~\cite{Imagetwin} and Proofig~\cite{Proofig}. Specifically, the images shown in Fig.~\ref{fig:showall} are submitted to the paid analysis services of both platforms, and both platforms report \textbf{0} AI-generated images\footnote{The reports are available for download at: \url{https://github.com/YoursEver/GANBlot}.}. Because these platforms provide commercial paid services, this experiment is intended as a limited supplementary screening of representative GAN-Blot examples rather than a large-scale platform benchmark. These results provide additional evidence that publisher-used screening platforms may not reliably identify GAN-Blot images under practical screening conditions. 

Finally, WesternBlot46K is intended to support the validation and development of WB forensic methods. Its controllable generation setting allows researchers to construct synthetic WB images with specified band structures and background appearances. The results in this section show that these generated images form challenging controlled cases for existing forensic tools. Developing a dedicated detector for experiment-inconsistent WB evidence requires additional expert annotation of generated WB images and remains an open problem for future study.

\section{Concluding Remarks}
\label{sec05:conclu}
We present GAN-Blot, a controllable Western blot (WB) image synthesis framework formulated through structure–style conditional generation. The proposed design enables independent control of band geometry and appearance characteristics without relying on predefined attribute annotations. Beyond the generator itself, we establish a benchmark for WB forensic research, including a large-scale synthetic WB dataset and evaluation protocols covering controllability, target-conditioned resynthesis, expert-based visual assessment, and forensic evaluation under existing detectors. Experimental results show that GAN-Blot produces structurally consistent and visually realistic WB images under diverse conditions. In addition, existing AI-generated image detectors, WB-specific synthetic-image detectors, and publisher-used image-screening platforms do not reliably identify GAN-Blot images and show limited reliability under WB-image forensic scenarios. These results show that GAN-Blot provides challenging controlled cases for validating and developing WB forensic methods, thereby supporting future research on scientific-image forensics and the protection of scientific integrity. 

\section*{Acknowledgments}
The authors would like to thank Dr. Cheng-Ting Chien, a Distinguished Research Fellow at the Institute of Cellular and Organismic Biology, Academia Sinica, Taiwan, for providing the source laboratory WB photos and sharing his expertise in WB-related experimental knowledge. The authors also thank Dr. Ya-Jen Cheng for helpful discussions on WB-related knowledge, and Dr. Yi-Chun Huang and Dr. Hsin-Ho Sung, two postdoctoral researchers in Dr. C.-T. Chien’s team, for preparing WB photos under various imaging configurations. The authors further thank Ms. Zih-Min Huang for helping rearrange and redraw Figs. 3 and 4. 

\bibliographystyle{IEEEtran}
\bibliography{refs}

\end{document}